%% file: egpaper_arxiv.tex
\documentclass[10pt,twocolumn,letterpaper]{article}

\usepackage{iccv}
\usepackage{times}
\usepackage{epsfig}
\usepackage{multirow}
\usepackage{graphicx}
\usepackage{amsmath}
\usepackage{amssymb}
\usepackage{booktabs}
\usepackage{color}
\usepackage{colortbl}
\usepackage{xcolor}
\usepackage{comment}
\definecolor{mygray1}{gray}{.9}
\definecolor{mygray2}{gray}{.7}
\definecolor{mygray1}{gray}{.9}
\definecolor{mygray2}{gray}{.7}

\newlength\savewidth\newcommand\shline{\noalign{\global\savewidth\arrayrulewidth
  \global\arrayrulewidth 1pt}\hline\noalign{\global\arrayrulewidth\savewidth}}

\newcommand{\rnum}[1]{\lowercase\expandafter{\romannumeral #1\relax}}

\usepackage[breaklinks=true,bookmarks=false]{hyperref}

\iccvfinalcopy 

\def\iccvPaperID{9047} 

\ificcvfinal\fi

\begin{document}

\title{DistillBEV: Boosting Multi-Camera 3D Object Detection with Cross-Modal Knowledge Distillation}

\author{
  Zeyu Wang${^{1,2*}}$ \quad Dingwen Li${^{1*}}$ \quad Chenxu Luo${^1}$ \quad Cihang Xie${^2}$ \quad Xiaodong Yang${^{1\dagger}}$ \\
  $^1$QCraft \quad $^2$UC Santa Cruz \\
}

\maketitle
\ificcvfinal\thispagestyle{empty}\fi

\def\thefootnote{*}\footnotetext{Equal contribution}
\def\thefootnote{$\dagger$}\footnotetext{Correspondence to \texttt{xiaodong@qcraft.ai}}

\begin{abstract}
3D perception based on the representations learned from multi-camera bird's-eye-view (BEV) is trending as cameras are cost-effective for mass production in autonomous driving industry. However, there exists a distinct performance gap between multi-camera BEV and LiDAR based 3D object detection. One key reason is that LiDAR captures accurate depth and other geometry measurements, while it is notoriously challenging to infer such 3D information from merely image input. In this work, we propose to boost the representation learning of a multi-camera BEV based student detector by training it to imitate the features of a well-trained LiDAR based teacher detector. We propose effective balancing strategy to enforce the student to focus on learning the crucial features from the teacher, and generalize knowledge transfer to multi-scale layers with temporal fusion. 
We conduct extensive evaluations on multiple representative models of multi-camera BEV. Experiments reveal that our approach renders significant improvement over the student models, leading to the state-of-the-art performance on the popular benchmark nuScenes. 
\end{abstract}

\section{Introduction}
\label{sec:intro}

Perceiving 3D environments is essential for autonomous driving as it is crucial for subsequent onboard modules from prediction~\cite{wayformer, wang2023prophnet} to planning~\cite{planning, li2023tip}. Although LiDAR based methods have achieved remarkable progress~\cite{lang2019pointpillars,luo2021simtrack,luo2021pillar,yin2021center}, there have recently been fast growing attentions to camera based approaches from both academia and industry~\cite{tesla,philion2020lift}. Compared to a sensor suite of LiDAR, the cost of cameras is typically 10 times cheaper, gaining a huge cost advantage for mass-production OEMs. Additionally, cameras are better suited to detect distant objects and recognize visual based road elements (e.g., traffic lights and signs). 

\begin{figure}
    \centering
    \includegraphics[width=\linewidth]{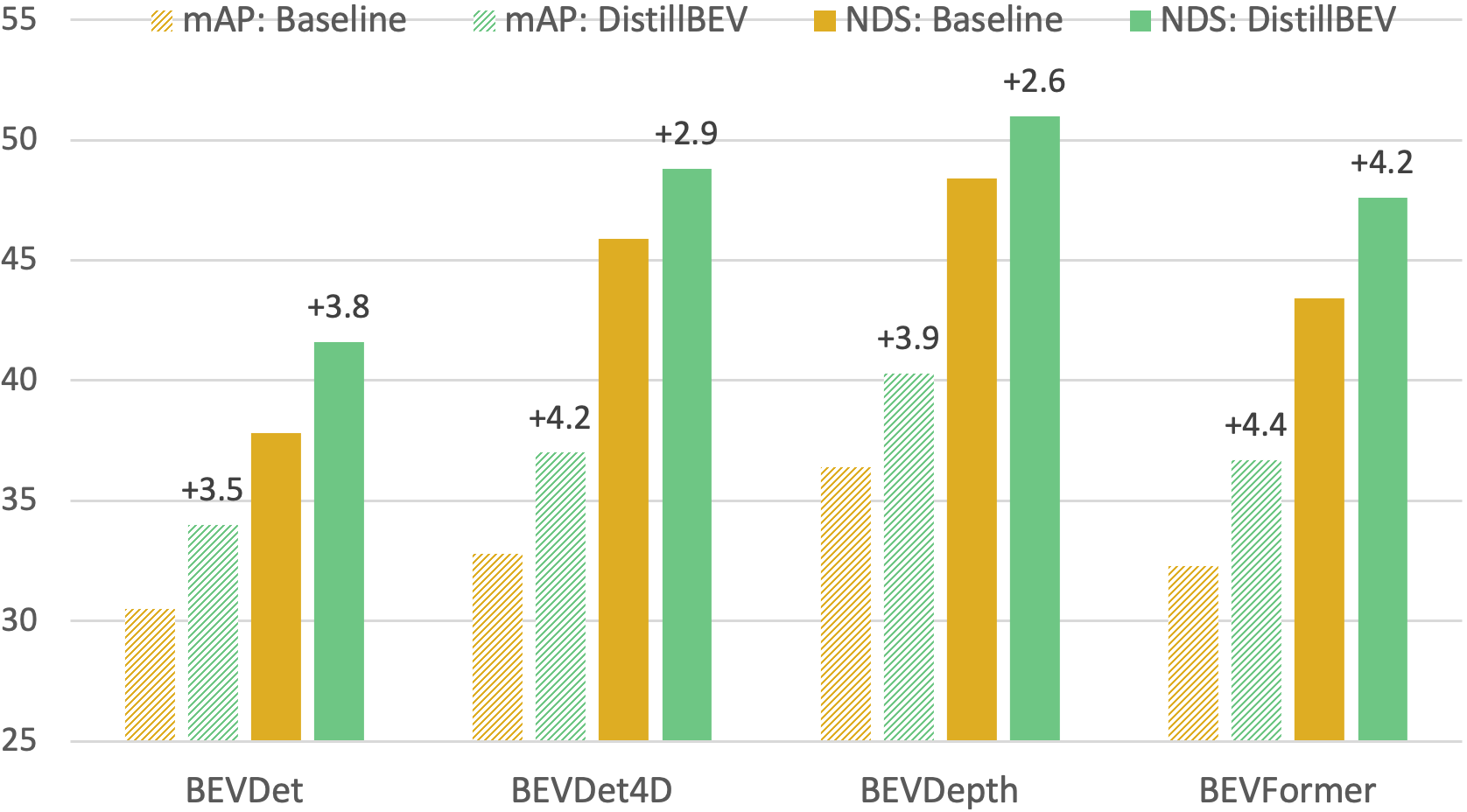}
    \caption{An overview of performance improvement in terms of mAP and NDS on the validation set of nuScenes. Enabled by the proposed cross-modal knowledge distillation method DistillBEV, a variety of multi-camera BEV based 3D object detectors achieve consistent and significant performance boost.}
    \label{fig:teaser}
\end{figure}

A straightforward solution for camera based 3D object detection is the monocular paradigm that has been broadly studied~\cite{ma2021delving,park2021pseudo,wang2021fcos3d}. However, such methods have to deal with surrounding cameras separately and require complex post-processing to fuse detections from multiple views. As an alternative, the bird's-eye-view (BEV) based framework is drawing extensive attentions to offer a holistic feature representation space from multi-camera images, and has made substantial improvement~\cite{huang2022bevdet4d,huang2021bevdet,li2022bevdepth,li2022bevformer}. BEV based framework owns the following inherent merits including (\rnum{1}) joint feature learning from multi-view images, (\rnum{2}) unified detection space without post fusion, (\rnum{3}) amenability for temporal fusion, and (\rnum{4}) convenient output representation for the downstream prediction and planning.      

Despite the advancement achieved in this field, a distinct performance gap remains between multi-camera BEV and LiDAR based 3D object detection. For instance, the leading multi-camera BEV based method is outperformed by its LiDAR counterpart over 15\% mAP and 10\% NDS on popular benchmark nuScenes~\cite{caesar2020nuscenes}. 
On the other hand, a data collection fleet can be equipped with both cameras and LiDAR, while the mass-produced vehicles can be LiDAR-free. 

In light of above observations, we present \textbf{DistillBEV}: a simple and effective cross-model knowledge distillation approach to bridge the feature learning gap between multi-camera BEV and LiDAR based detectors. 
Our strategy toward achieving this goal is to align the corresponding features learned from images and point clouds. Given merely images as input, the multi-camera BEV based detector (student) is guided to imitate the features extracted from point clouds by a well-trained LiDAR based detector (teacher). We argue that the accurate 3D geometric cues such as depth and shape as well as how such cues are represented in the point cloud features provide valuable guidance to the training process of student model. Moreover, we emphasize that our approach incurs no extra computation cost during inference as the teacher model involves in training only.  

Due to the notable discrepancy between the two modalities, the cross-modal knowledge distillation is extremely challenging. 
Compared to camera images, LiDAR scans are inherently sparse, and the majority of 3D space is empty. 
Even if in the occupied space, background (e.g., building and vegetation) dominates, and meanwhile, objects of interest (e.g., bus and pedestrian) are with largely varied sizes.  
It is thus nontrivial to locate the informative regions to make the knowledge transfer more focused, and balance the distillation importance assigned to different foreground objects. Furthermore, the teacher and student models are respectively developed in their specific domains, leading to disparate network architectures. It is also demanding to generalize the cross-modal distillation to adapt to various combinations of different teacher and student networks.  

To tackle these challenges for DistillBEV, we first exploit region decomposition to partition a feature map into true positive, false positive, true negative and false negative regions, which elaborately decouple foreground and background in the cross-modal distillation. Based on this decomposition, we introduce adaptive scaling to balance the significantly varied box sizes when objects are presented in BEV. We further utilize spatial attention to encourage the student model to mimic the attention pattern produced by the teacher model, so as to focus on the crucial features for effective knowledge transfer. We then extend the distillation to multi-scale layers to achieve thorough feature alignment between teacher and student. Finally, we incorporate temporal information for both teacher and student models in BEV, thus enabling the distillation with temporal fusion readily. Thanks to the proposed generalizable design, our approach is flexible to be applied to various combinations of teacher and student detectors. As illustrated in Figure~\ref{fig:teaser}, DistillBEV consistently and remarkably improves multiple representative student models.    

Our main contributions are summarized as follows. First, we present the cross-modal distillation in BEV, which naturally suits for knowledge transfer between LiDAR and multi-camera BEV based detectors. Second, we propose the effective balancing design to enable the student to focus on learning crucial features of the teacher with multiple scales and temporal fusion. Third, our approach achieves superior performance, and more importantly, obtains consistent and considerable improvement on various teacher-student combinations. Our code and model will be made available at \url{https://github.com/qcraftai/distill-bev}.   

\section{Related Work}

\noindent\textbf{Camera based 3D Object Detection.} 
A large family of the methods in this field are based on the monocular paradigm, such as FCOS3D~\cite{wang2021fcos3d} and DD3D~\cite{park2021pseudo}, which resemble 2D object detection. Recently, the multi-camera BEV based framework is trending thanks to its inherent merits. In this framework, the view transformation module plays a fundamental role to convert multi-view image features to BEV. 
Some methods employ inverse perspective mapping~\cite{bev-ipm} or multilayer perceptron~\cite{bev-mlp} to perform the translation from perspective view to BEV. 
In~\cite{philion2020lift} LSS is introduced to lift image features by corresponding bin-wise depth distribution. This line of research includes a series of works such as BEVDet~\cite{huang2021bevdet}, BEVDet4D~\cite{huang2022bevdet4d}, and BEVDepth~\cite{li2022bevdepth}. 
BEVFormer is proposed in~\cite{li2022bevformer} to utilize cross-attention to look up and aggregate image features across cameras. In addition, BEV representations provide a more desirable connection of scene features at multiple timestamps to improve object detection and motion state estimation. BEVDet4D and BEVDepth fuse previous and current features using spatial alignment, and BEVFormer performs fusion through temporal attention in a soft way.

\noindent\textbf{LiDAR based 3D Object Detection.}
Since most methods of this field apply voxelization to transform irregular point clouds to regular grids such as pillars or voxels, it is natural to extract features in BEV. VoxelNet~\cite{zhou2018voxelnet} applies 3D convolutions on the aggregated point features inside each voxel. SECOND~\cite{yan2018second} takes advantage of sparse 3D convolutions to improve computation efficiency. PointPillars~\cite{lang2019pointpillars} proposes to collapse height dimension and use 2D convolutions to further reduce inference latency. CenterPoint~\cite{yin2021center} is a popular anchor-free method that represents objects as points. PillarNeXt~\cite{li2023pillar} shows that the pillar based models with modernized designs in architecture and training outperforms the voxel counterpart in both accuracy and latency. Fusing multiple sensors is also widely used to enhance the detection performance.  
MVP~\cite{yin2021multimodal} is a sensor-fusion version of CenterPoint enhanced by image virtual points.  

\begin{figure*}
    \centering
    \includegraphics[width=\linewidth]{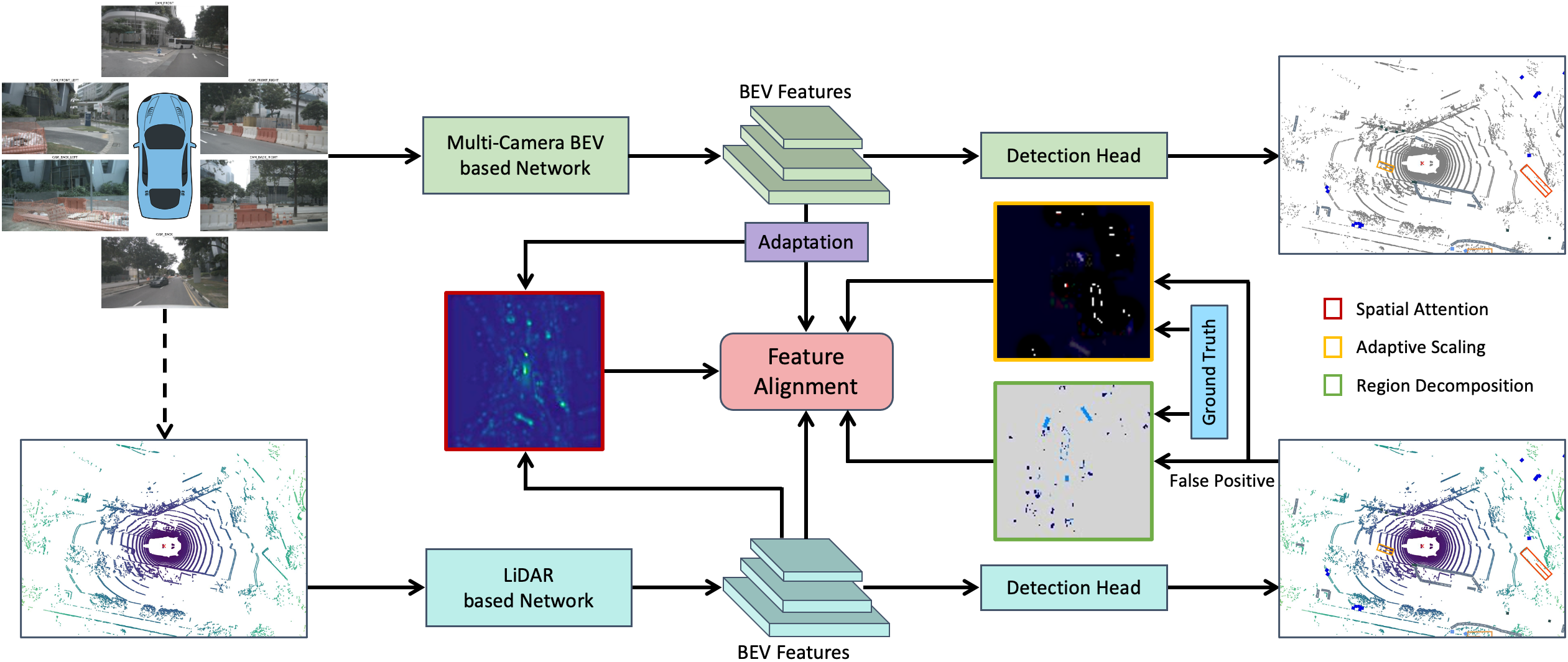}
    \vspace{-2pt}
    \caption{A schematic overview of the proposed cross-modal distillation approach DistillBEV. We aim to bridge the representation learning gap between multi-camera BEV (student) and LiDAR (teacher) based detectors by guiding the former to imitate the features extracted by the latter. We introduce the balancing strategy including region decomposition, adaptive scaling and spatial attention to encourage the student to focus on the crucial features to align with the teacher. Our teacher model can be based on either LiDAR or camera-LiDAR fusion (indicated by the dashed line). Note the bottom branch of LiDAR related components are removed after training.}
    \label{fig:pipeline}
\end{figure*}

\noindent\textbf{Knowledge Distillation.}
This technique is originally proposed for network compression by transferring the information from a larger teacher model to a compact student model~\cite{hinton2015distilling}. Most methods in the field are initially designed for image classification, but hardly make improvement for image object detection. Some recent methods have successfully adapted knowledge distillation for 2D object detection~\cite{dai2021general,kang2021instance,yang2022focal,zhang2021improve}. Nevertheless, the research on distillation for 3D object detection has been less explored, in particular when the teacher and student models come from different modalities~\cite{chong2022monodistill, li2022unifying}. 
The most related work to our proposed approach is~\cite{chen2022bevdistill}, which introduces a dense foreground-guided feature imitation and a sparse instance-wise distillation to transfer spatial knowledge from LiDAR to multi-camera 3D object detection. Compared to this method, our approach achieves more fine-grained distillation through introducing region decomposition and adaptive scaling. Additionally, our design accommodates multi-scale distillation, which enhances cross-modal knowledge transfer at different feature abstractions. Experimental results also confirm the superiority of our approach.           


\section{Method}
As illustrated in Figure~\ref{fig:pipeline}, to tackle the challenges for cross-modal knowledge transfer, DistillBEV involves formulating region decomposition mask, adaptive scaling factor and spatial attention map, as well as making extensions to multi-scale layers and temporal fusion. 

\subsection{Region Decomposition}
It has been well acknowledged in 2D object detection that simply performing feature alignment between teacher and student hardly makes improvement due to the imbalance between foreground and background regions~\cite{dai2021general,kang2021instance,yang2022focal,zhang2021improve}. However, this phenomenon is even more severe in 3D object detection as the vast majority of 3D space is unoccupied. Our statistics on BEV feature maps finds that less than 30\% pixels are non-empty on average, and among them, only a small fraction contains objects that we are interested in. To perform effective knowledge transfer, we introduce the region decomposition to guide the student to focus on crucial regions rather than treating all regions equally. Specifically, we partition a feature map into four types: true positive (TP), false positive (FP), true negative (TN) and false negative (FN). Accordingly, we define a region decomposition mask $M$:   
\begin{equation}
M_{i, j}= \begin{cases}
1, & \text { if }(i, j) \in \text{TP or FN} \\
\eta, & \text { if }(i, j) \in \text{FP} \\ 
0, & \text { if }(i, j) \in \text{TN} 
\end{cases}
\label{Eq:FG Mask}
\end{equation}
where $i, j$ are the coordinates on a feature map, and $\eta$ controls the relative importance of pixels in FP regions. 

This decomposition gives our approach the flexibility in assigning varied importance on different regions. It is straightforward to take into account the regions covered by ground truth boxes (i.e., the union of TP and FN regions), which exactly convey the features from foreground objects. However, we also treat FP regions differently from TN regions. When the teacher model generates high activations on some regions even though they are FP (e.g., a pole is misdetected as a pedestrian), encouraging the student model to mimic such feature responses can be still beneficial for the the overall 3D geometry learning.
FP regions can be found by thresholding the confidence heatmaps produced by the teacher detector and ground truth labels: 
\begin{equation}
\begin{aligned}
    \text{FP} = \left(H^t > \gamma\right) \&\left(H^g < \gamma\right),
\end{aligned}
\label{Eq:FP region}
\end{equation}
where $H^t$ and $H^g$ correspond to the heatmaps obtained from the teacher model and ground truth, respectively, and $\gamma$ is a hyper-parameter for heatmap thresholding.

\subsection{Adaptive Scaling}
Another challenge to distill knowledge from teacher to student in BEV is the great span of various object sizes. For instance, a bus is dozens of times the size of a pedestrian from bird's-eye-view. Moreover, background such as walls and plants overwhelm the non-empty regions. Thus, background stuff and giant foreground objects would dominate the distillation loss since substantially more features come from them. It is desirable to reflect  different-sized objects or classes with similar contributions in the distillation loss. An adaptive scaling factor is introduced to achieve this goal:
\begin{equation}
S_{i, j}= 
\begin{cases}
\frac{1}{\sqrt{H_k W_k}}, & \text { if }(i, j) \in O_k \\ 
\frac{1}{N_{\text{FP}}}, & \text { if }(i, j) \in \text{FP} \\ 
\frac{1}{N_{\text{TN}}}, & \text { if }(i, j) \in \text {TN}
\end{cases}
\label{Eq:Scale Mask}
\end{equation}
where $O_k$ is the $k$-th ground truth object (TP or FN) with bounding box length $H_k$ and width $W_k$ in BEV, $N_{\text{FP}}$ and $N_{\text{TN}}$ denote the number of pixels falling in the FP and TN regions, respectively.

\subsection{Spatial Attention}
Attention maps~\cite{hu2018squeeze,yang2022focal,zhang2021improve} have been applied to various tasks and architectures to improve representation learning by focusing on the important features as well as suppressing the insignificant ones. Here we adopt a spatial attention map based on the extracted features of both teacher and student to further select more informative features to concentrate on. A spatial attention map is constructed by: 
\begin{equation}
\mathcal{P}(F)_{i,j} = \frac{1}{C} \sum_{c=1}^C\left|F_{c,i,j}\right|,
\end{equation}
\begin{equation}
\mathcal{N}(F) = HW \operatorname{softmax}\left(\mathcal{P}(F) / \tau\right),
\end{equation}
where $F \in \mathbb{R}^{C \times H \times W}$ is a feature map, $\mathcal{P}(F) \in \mathbb{R}^{H \times W}$ denotes the average pooling result of the absolute values of $F$ along the channel dimension, $\mathcal{N}(F) \in \mathbb{R}^{H \times W}$ is the normalized attention by softmax over all spatial locations, and $
\tau$ is a temperature to adjust the distribution entropy. We obtain the final spatial attention map by considering the feature maps from both teacher $F^t$ and student $F^s$:   
\begin{equation}
A = (\mathcal{N}(F^t) + \mathcal{N}(\Tilde{F}^s)) / 2, \hspace{10pt} \Tilde{F}^s = \mathcal{G}(F^s),
\end{equation}
where $\mathcal{G}$ is an adaptation module to map $F^s$ to $\Tilde{F}^s$ that is with the same size as $F^t$. 
More details of the design choices about $\mathcal{G}$ are discussed in Sections \ref{sec:multi-scale} and \ref{sec:ablation study}. 

\subsection{Multi-Scale Distillation} 
\label{sec:multi-scale}
It is commonly believed that layers of different depth in a network encodes varying feature abstractions~\cite{kriegeskorte2015deep}. One successful application is the feature pyramid network~\cite{lin2017feature}, which combines features from different levels to better detect objects of various sizes~\cite{li2023pillar,tian2019fcos,yin2021center}. 
To realize comprehensive alignment between teacher and student, we adopt this idea to perform feature distillation at multiple scales for the models based on CNNs. 
However, the teacher and student networks are separately designed with different architectures, making it nontrivial to find intermediate feature correspondence. For instance, the BEV feature maps in teacher are usually 2$\times$ or 4$\times$ the size of those in student. Naively aligning features of the same resolution results in incompatibility of the feature abstraction level. Hence, we introduce a lightweight adaptation module $\mathcal{G}$ consisting of upsampling and projection layers to map the student feature before aligning with the teacher feature at the similar level. We also find that feature imitation at early layers is detrimental to the distillation, which is because the representation discrepancy caused by the modality gap between point clouds and images remains substantial at the early stage.   
Note we identify and utilize the FP regions only at the last encoding layer of BEV (i.e., the pre-head feature). We find this setting works the best presumably because FP regions are better expressed by high-level semantic features emerging at the last layer. See more architecture details in the supplementary material. 

\subsection{Distillation Loss}
We train a student network with the original loss including classification and regression as well as the overall distillation loss, the related terms of which are summarized as follows. We first define the feature imitation loss at the $n$-th distillation layer between teacher $F^{t(n)}$ and student $F^{s(n)}$: 
\begin{align}
\begin{split}
&L_{\text{feat}} = \alpha \sum_{c=1}^C \sum_{i=1}^H \sum_{j=1}^W M_{i, j} S_{i, j} A_{i, j} \left(F_{c, i, j}^{t(n)}-\Tilde{F}^{s(n)}_{c, i, j}\right)^2 \\
&+\beta \sum_{c=1}^C \sum_{i=1}^H \sum_{j=1}^W\ \overline{M}_{i, j} S_{i, j} A_{i, j} \left(F_{c, i, j}^{t(n)}-\Tilde{F}^{s(n)}_{c, i, j}\right)^2,
\end{split}
\label{eq:imitation-loss}
\end{align}
where $\Tilde{F}^{s(n)} = \mathcal{G}(F^{s(n)})$, $\overline{M}$ is the logical complement of the region decomposition mask $M$, $S$ denotes the adaptive scaling factor, $A$ is the spatial attention map, and $\alpha$ and $\beta$ are the hyper-parameters to weight the two terms. 

Furthermore, we exploit an attention imitation loss to enforce the student to learn to generate an attention pattern similar to the teacher, and therefore focus on the spatial locations that the teacher network considers more crucial:
\begin{equation}
L_{\text{attn}} = \sum_{i=1}^H \sum_{j=1}^W \left|\mathcal{P}(F^{t(n)})_{i,j} - \mathcal{P}(\Tilde{F}^{s(n)})_{i,j}\right|.
\label{eq:attention-loss}
\end{equation}

In summary, the overall distillation objective function is the sum of feature imitation loss (\ref{eq:imitation-loss}) and attention imitation loss (\ref{eq:attention-loss}) at multiple scales:   
\begin{equation}
\hspace{-5pt} L_{\text{dist}}=\sum_{n=1}^N L_{\text{feat}}(F^{t(n)}, F^{s(n)})+ \lambda L_{\text{attn}}(F^{t(n)}, F^{s(n)}),
\end{equation}
where $N$ is the number of selected layers to perform distillation, and $\lambda$ controls the relative importance between the two loss functions.

\subsection{Distillation with Temporal Fusion}
One desirable property of the representations in multi-camera BEV is to facilitate fusing of features from multiple timestamps. Methods~\cite{huang2022bevdet4d,li2022bevdepth,li2022bevformer} developed with temporal fusion greatly improve 3D object detection and motion estimation by leveraging important dynamic cues. As for LiDAR based models, it is a common practice to fuse multiple point clouds by directly transforming the past sweeps to the current coordinate frame through ego-motion compensation, and a relative timestamp is added to the measurements of each point~\cite{yin2021center}. Thus, it is natural to conduct temporal knowledge transfer in our approach as the teacher can be readily compatible with the student making use of temporal information. In practice, we employ a unified teacher model for both single-frame and multi-frame based student models to perform distillation with temporal fusion.

\section{Experiments}
In this section, we describe the evaluation setup including dataset, metrics, and implementation details. A variety of ablation study and related analysis are conducted for in-depth understanding of each individual component in our approach. We report extensive comparisons with the state-of-the-art methods on the popular benchmark. 

\subsection{Experimental Setup}
\noindent\textbf{Dataset and Metrics.} We evaluate our approach on the large-scale autonomous driving benchmark nuScenes~\cite{caesar2020nuscenes}. This dataset consists of 1,000 scenes of roughly 20 seconds each, captured by a 32-beam LiDAR and 6 cameras at the frequency of 20Hz and 10Hz. There are 10 classes in total for 3D object detection, and the annotations are provided at 2Hz. Following the standard evaluation split, 700, 150 and 150 scenes are respectively used for training, validation and test. We follow the official evaluation metrics including mean average precision (mAP) and nuScenes detection score (NDS) as the main metrics. We also use mATE, mASE, mAOE, mAVE and mAAE to measure translation, scale, orientation, velocity and attribute related errors.

\noindent\textbf{Teacher and Student Models.} To validate the generalizability of our approach, we consider various teacher and student models. We employ the popular CenterPoint~\cite{yin2021center} or its sensor-fusion version MVP~\cite{yin2021multimodal} as the teacher model. More details are in the supplementary material. As for the student model, we choose BEVDet~\cite{huang2021bevdet}, BEVDet4D~\cite{huang2022bevdet4d}, BEVDepth~\cite{li2022bevdepth}, and BEVFormer~\cite{li2022bevformer} as the representative student models, which represent a broad range of student models from CNNs to Transformers, as well as from the basic version to the temporal (``4D'' to incorporate temporal fusion) and spatial (``Depth'' to enhance trustworthy depth estimation) extensions. These models together form 8 different teacher-student combinations.

\begin{table*}[t]
\centering
\small
\begin{tabular}{lc|cc|cc|ccccc}
\shline
Teacher &Mode &Student &Mode &mAP$\uparrow$ &NDS$\uparrow$ &mATE$\downarrow$ &mASE$\downarrow$ &mAOE$\downarrow$ &mAVE$\downarrow$ &mAAE$\downarrow$ \\ \shline
- &- &BEVDet &C &30.5 &37.8 &72.1 &27.9 &57.9 &91.4 &25.0 \\
CenterPoint &L &BEVDet &C &32.7 &40.7 &70.9 &26.5 &56.5 &81.2 &21.0 \\
MVP &L\&C &BEVDet &C &\textbf{34.0} &\textbf{41.6} &70.4 &26.6 &55.6 &81.5 &20.1 \\ \hline
- &- &BEVDet4D &C &32.8 &45.9 &69.5 &27.9 &50.8 &36.5 &20.6 \\
CenterPoint &L &BEVDet4D &C &36.3 &48.4 &66.6 &26.8 &49.8 &34.9 &19.9 \\
MVP &L\&C &BEVDet4D &C &\textbf{37.0} &\textbf{48.8} &67.6 &26.8 &46.1 &36.8 &20.0 \\ \hline
- &- &BEVDepth &C &36.4 &48.4 &64.9 &27.3 &49.8 &34.9 &20.7 \\
CenterPoint &L &BEVDepth &C &38.9 &49.8 &63.0 &26.7 &50.4 &36.0 &20.2 \\
MVP &L\&C &BEVDepth &C &\textbf{40.3} &\textbf{51.0} & 62.3 & 26.6 & 46.4 & 35.7 & 20.7 \\ \hline
- &- &BEVFormer &C & 32.3 & 43.4 & 79.6 & 28.3 & 53.1 & 46.0 & 21.4 \\
CenterPoint &L &BEVFormer &C & 35.6 & 47.0 & 71.6 & 27.3 & 49.1 & 39.7 & 19.8 \\
MVP &L\&C &BEVFormer &C & \textbf{36.7} & \textbf{47.6} & 72.1 & 27.5 & 50.6 & 37.6 & 20.0 \\\shline
\end{tabular}
\vspace{5pt}
\caption{Comparison of our approach using various combinations of teacher and student models on the validation set of nuScenes. ``C'' and ``L'' indicate the modality of camera and LiDAR, respectively.}
\vspace{-9pt}
\label{tab:val set results}
\end{table*}

\begin{table}[t]
\centering
\resizebox{\linewidth}{!}{
\begin{tabular}{lcc|cc}
\shline
Method &Backbone &Mode &mAP &NDS \\ \shline
CenterPoint~\cite{yin2021center} &- &L &56.4 &64.8 \\
MVP~\cite{yin2021multimodal} &- &L\&C &67.1 &78.0 \\
\hline
FCOS3D~\cite{wang2021fcos3d} &R101 &C &34.3 &41.5 \\
PETR~\cite{liu2022petr} &R101 &C &35.7 &42.1 \\
DETR3D~\cite{wang2022detr3d} &R101 &C &34.6 &42.5 \\
BEVFormer~\cite{li2022bevformer} &R50 &C &32.3 &43.4 \\
BEVFormer~\cite{li2022bevformer} &R101 &C &41.6 &51.7 \\
BEVDepth~\cite{li2022bevdepth} &R50 &C &35.1 &47.5 \\
BEVDepth~\cite{li2022bevdepth} &R101 &C &41.2 &53.5 \\
 \hline
Set2Set~\cite{wang2021object} &R50 &C &37.5 &47.9 \\
FitNet~\cite{romero2014fitnets} &R50 &C &37.3 &48.0 \\
MonoDistill~\cite{chong2022monodistill} &R50 &C &39.0 &49.5 \\
UVTR~\cite{li2022unifying} &R50 &C &39.4 &50.1 \\
BEVDistill~\cite{chen2022bevdistill} &R50 &C &40.7 &51.5 \\ 
BEVDistill~\cite{chen2022bevdistill} &R101 &C &41.7 &52.4 \\  
\hline
Ours (BEVFormer) &R50 &C &36.7 &47.6 \\
Ours (BEVFormer) &R101 &C &44.6 &54.5 \\
Ours (BEVDepth) &R50 &C &40.3 &51.0 \\
Ours (BEVDepth) &R101 &C &\textbf{45.0} &\textbf{54.7} \\ \shline
\end{tabular}
}
\vspace{1pt}
\caption{Comparison on the validation set of nuScenes. Groups 1-4 correspond to the teacher models, the camera based works, the distillation based methods, and our proposed approach.  
}
\label{tab:val sota comparison}
\end{table}

\noindent\textbf{Implementation Details.} 
We implement our approach in PyTorch~\cite{paszke2019pytorch}, 
and train the networks by using 8 NVIDIA Tesla V100 GPUs with the batch size of 64. 
AdamW~\cite{loshchilov2018decoupled} is adopted as the optimizer with a cosine-scheduled learning rate of 2e-4. 
All models are trained for 24 epochs with the strategy of CBGS~\cite{zhu2019class}. 
Following~\cite{huang2021bevdet,li2022bevdepth}, data augmentations are applied in both image and BEV spaces.  
We follow the standard evaluation protocol to set the detection range to [-51.2m, 51.2m]$\times$[-51.2m, 51.2m]. 
ResNet-50~\cite{he2016deep} pre-trained on ImageNet-1K is used as image backbone and image size is processed to 256$\times$704, unless otherwise specified. 
We adopt the common inheriting practice~\cite{kang2021instance} to initialize the detection head of student by the parameters of teacher for faster convergence. 
More details can be found in the supplementary material. 

\subsection{Comparison with State-of-the-Art Methods}
\label{subsec:Main Results}
We start from comparing the student performance before and after distillation on the validation set of nuScenes. As shown in Table~\ref{tab:val set results} and Figure~\ref{fig:teaser}, DistillBEV using different teacher models considerably and consistently boosts the four representative students over various metrics. In particular, the most significant performance gains are obtained on BEVFormer, i.e., 4.4\% mAP and 4.2\% NDS. For the leading algorithm BEVDepth, we also observe the improvement of 3.9\% mAP and 2.6\% NDS, revealing that our distillation effect is not diminishing with the stronger student. 

Taking a closer look into other metrics, we observe that DistillBEV largely improves mAVE of the single-frame based model (i.e., BEVDet). We attribute this to the temporal fusion nature in the teacher models, which effectively transfer temporal knowledge to the distilled students via our approach. 
And the trend of mAAE follows mAVE as this attribute is predicted based on the velocity estimation. In general, there exits clear improvement in mATE and mAOE, thanks to the accurate depth and geometric cues encoded in point clouds. It is found that mASE is only slightly improved as the students already estimate object scale reasonably well by using visual information alone.

We next compare DistillBEV with the state-of-the-art methods on the validation set and test set of nuScenes. Table~\ref{tab:val sota comparison} shows that, on the validation set, our approach outperforms the camera based methods (2nd group) by a clear margin with the same backbone settings. Comparing with other knowledge distillation based methods (3rd group), our approach also performs better than various competing algorithms.  
As for the test set, we follow~\cite{li2022bevdepth} to increase the input image size to 640$\times$1600 and double the grid size to 256$\times$256 in BEV, and the pre-training on ImageNet-22K is further applied. As compared in Table~\ref{tab:test sota comparison}, our distillation improves the baseline by 3.6\% mAP and 2.2\% NDS, and achieves superior performance without bells and whistles, such as model ensembling and test-time augmentation. 

\begin{table}[t]
\centering
\resizebox{\linewidth}{!}{
\begin{tabular}{lcc|cc}
\shline
Method &Backbone &Mode &mAP &NDS \\ \shline
CenterPoint~\cite{yin2021center} &- &L &58.0 &65.5 \\
MVP~\cite{yin2021multimodal} &- &L\&C &66.4 &70.5 \\
\hline
FCOS3D~\cite{wang2021fcos3d} &R101 &C &35.8 &42.8 \\
BEVDet~\cite{huang2021bevdet} &Swin-B &C &39.8 &46.3 \\
DD3D~\cite{park2021pseudo} &VoV-99 &C &41.8 &47.7 \\
DETR3D~\cite{wang2022detr3d} &VoV-99 &C &41.2 &47.9 \\
PETR~\cite{liu2022petr} &VoV-99 &C &44.1 &50.4 \\
BEVDet4D~\cite{huang2021bevdet} &Swin-B &C &45.1 &56.9 \\
BEVFormer~\cite{li2022bevformer} &VoV-99 &C &48.1 &56.9 \\
BEVDistill~\cite{chen2022bevdistill} &ConvNeXt-B &C &49.8 &59.4 \\
BEVDepth~\cite{li2022bevdepth} &VoV-99 &C &50.3 &60.0 \\
BEVDepth~\cite{li2022bevdepth} &Swin-B &C &48.9 &59.0 \\
\hline
Ours (BEVDepth) &Swin-B &C &\textbf{52.5} &\textbf{61.2} \\ \shline
\end{tabular}
}
\vspace{1pt}
\caption{Comparison on the test set of nuScenes. Groups 1-3 are the teacher models, the state-of-the-art methods, and our proposed approach. Note that VoV-99~\cite{lee2019energy} is pre-trained on depth prediction with extra data of 15M images and paired point clouds~\cite{park2021pseudo}.
}   
\label{tab:test sota comparison}
\end{table}

\begin{table*}[t]
\centering

\begin{subtable}[h]{0.34\textwidth}
\centering
\small
    \begin{tabular}{c|c|cc}
    \shline
    Model  &Threshold &mAP &NDS \\
    \shline
    \multirow{3}{*}{Teacher} &0.05 & 36.9 & 48.4 \\
    & \cellcolor{mygray1} 0.1 & \cellcolor{mygray1} 37.8 & \cellcolor{mygray1} 49.0 \\
     &0.3 & 37.6 & 49.0 \\ 
     \hline
     Student &0.1 & 37.5 & 48.7\\
    \shline
    \end{tabular}
    \caption{Evaluation of different FP region mining principles by teacher and student under various thresholds.}
    \label{tab:fp selection}
\end{subtable}
\hfill
\begin{subtable}[h]{0.27\textwidth}
\centering
\small
\begin{tabular}{c|cc}
    \shline
    FP  &mAP &NDS \\
    \shline
    \rowcolor{mygray1} Pre-Head (H) & 37.8 & 49.0 \\ 
     \hline
     All (H-B2-B1) & 37.4 & 48.6 \\
    \shline
    \end{tabular}
    \caption{Evaluation of the impacts of FP region decomposition applied with different combinations of distillation layers.}
    \label{tab:fp selection position}
\end{subtable}
\hfill
\begin{subtable}[h]{0.37\textwidth}
\centering
\small
    \begin{tabular}{cccc|cc}
    \shline
    B0 &B1 &B2 &H &mAP &NDS \\ \shline
    & & & & 32.8 & 45.9 \\ \hline
    & & &\checkmark & 34.9 & 46.2 \\ 
    & &\checkmark &\checkmark & 36.1 & 47.3 \\
    \rowcolor{mygray1} &\checkmark &\checkmark &\checkmark & 36.8 & 48.4 \\
    \checkmark &\checkmark &\checkmark &\checkmark & 34.4 & 46.6\\ 
    \shline
    \end{tabular}
    \caption{Evaluation of the distillation effects at different layers.}
    \label{tab:location}
\end{subtable}

\begin{subtable}[h]{0.51\textwidth}
\centering
\small
\begin{tabular}{c|cccc|cc}
\shline
Method &FG\&BG &Attention &Scaling  &FP &mAP &NDS \\ \shline
Baseline & & & & & 32.8 & 45.9 \\ \hline
\multirow{5}{*}{Distill} & & & & & 32.9 & 45.6 \\
&\checkmark & & & & 33.4 & 46.3 \\
&\checkmark &\checkmark & & & 34.6 & 46.5 \\
&\checkmark &\checkmark &\checkmark & & 36.8 & 48.4 \\
&\cellcolor{mygray1}\checkmark &\cellcolor{mygray1}\checkmark &\cellcolor{mygray1}\checkmark &\cellcolor{mygray1}\checkmark & \cellcolor{mygray1}37.8 & \cellcolor{mygray1}49.0 \\ \shline
\end{tabular}
\caption{Comparison of our approach using different combinations of the proposed region decomposition mask including regions from FG\&BG and FP, spatial attention map, as well as adaptive scaling factor.}
\label{tab:ablation}
\end{subtable}
\hfill
\begin{subtable}[h]{0.44\textwidth}
\centering
\small
    \begin{tabular}{c|c|cc}
    \shline
    Layer &Adaptation Module  &mAP &NDS \\
    \shline
    \multirow{3}{*}{\shortstack{Pre-Head\\(H)}} & Linear & 37.6 & 48.8 \\
    & \cellcolor{mygray1} 2$\times$Block & \cellcolor{mygray1} 37.8 & \cellcolor{mygray1} 49.0 \\
    &3$\times$Block & 37.0 & 48.6 \\ \hline
    \multirow{4}{*}{\shortstack{Intermediate\\(B2-B1)}} &Linear & 37.3 & 48.9 \\
    &Upsample-2$\times$Block & 37.0 & 48.8 \\
    & \cellcolor{mygray1} Upsample-3$\times$Block & \cellcolor{mygray1} 37.8 & \cellcolor{mygray1} 49.0 \\
    &Downsample-3$\times$Block & 37.3 & 48.1 \\
    \shline
    \end{tabular}
    \caption{Comparison of different design choices of the adaptation module at different distillation layers. A linear projection is a 1$\times$1Conv and each block consists of 1$\times$1Conv-BN-ReLU.}
    \label{tab:adaptation network}
\end{subtable}
\label{table:design choices}
\caption{A set of ablation studies and related design choices on the validation set of nuScenes.}
\end{table*}

\subsection{Ablation Study}
\label{sec:ablation study}
We perform extensive experiments to analyze and understand the proposed components and related design choices in DistillBEV. Unless otherwise mentioned, we disable the head inheriting, and exemplify the teacher and student detectors using MVP and BEVDet4D, respectively. 

\noindent\textbf{Contribution of Individual Component.}
Here we study the contribution of each individual component in our design through a set of combinatorial experiments. Starting from the basic feature imitation at multiple scales, we gradually add foreground and background region decomposition (no FP), spatial attention, adaptive scaling, and FP region decomposition. As shown in Table~\ref{tab:ablation}, the naive feature imitation barely improves the baseline (32.8\% vs 32.9\% mAP, 45.9\% vs 45.6\% NDS), indicating that the basic feature alignment is inadequate for cross-modal distillation in BEV. In the proposed approach, each component makes decent improvement upon the basic feature imitation, and in the full model, these components collectively boost the baseline result by 5.0\% mAP and 3.1\% NDS.

\noindent\textbf{Where to Distill.}
Next we study the impact of distilling at different layers between teacher and student in DistillBEV. As mentioned in Section~\ref{sec:multi-scale}, FP regions are only used on the pre-head feature map. So we disable the FP region decomposition for a fair comparison among different levels.    
For the cross-modal distillation, the teacher and student networks are separately developed with distinct architectures, and the modality gap further complicates the problem. This differs from 2D object detection distillation~\cite{kang2021instance,zhang2021improve}, where the teacher and student models can adopt similar architectures and both take images as input. 
To alleviate this issue, we align features at similar abstraction level via the adaptation module.
We use H to indicate the pre-head layer, and B2-B0 to denote its three preceding intermediate layers
of the BEV encoders in both teacher and student.
As shown in Table~\ref{tab:location}, our multi-scale distillation shows a clear advantage over distilling at the pre-head layer alone (36.8\% vs 34.9\% mAP, 48.4\% vs 46.2\% NDS). It is also noteworthy that aligning features from an early layer (B0) is negative, presumably because the representation gap between point clouds and images remains large at the early stage.

\begin{figure*}[h]
    \centering
    \includegraphics[width=0.96\linewidth]{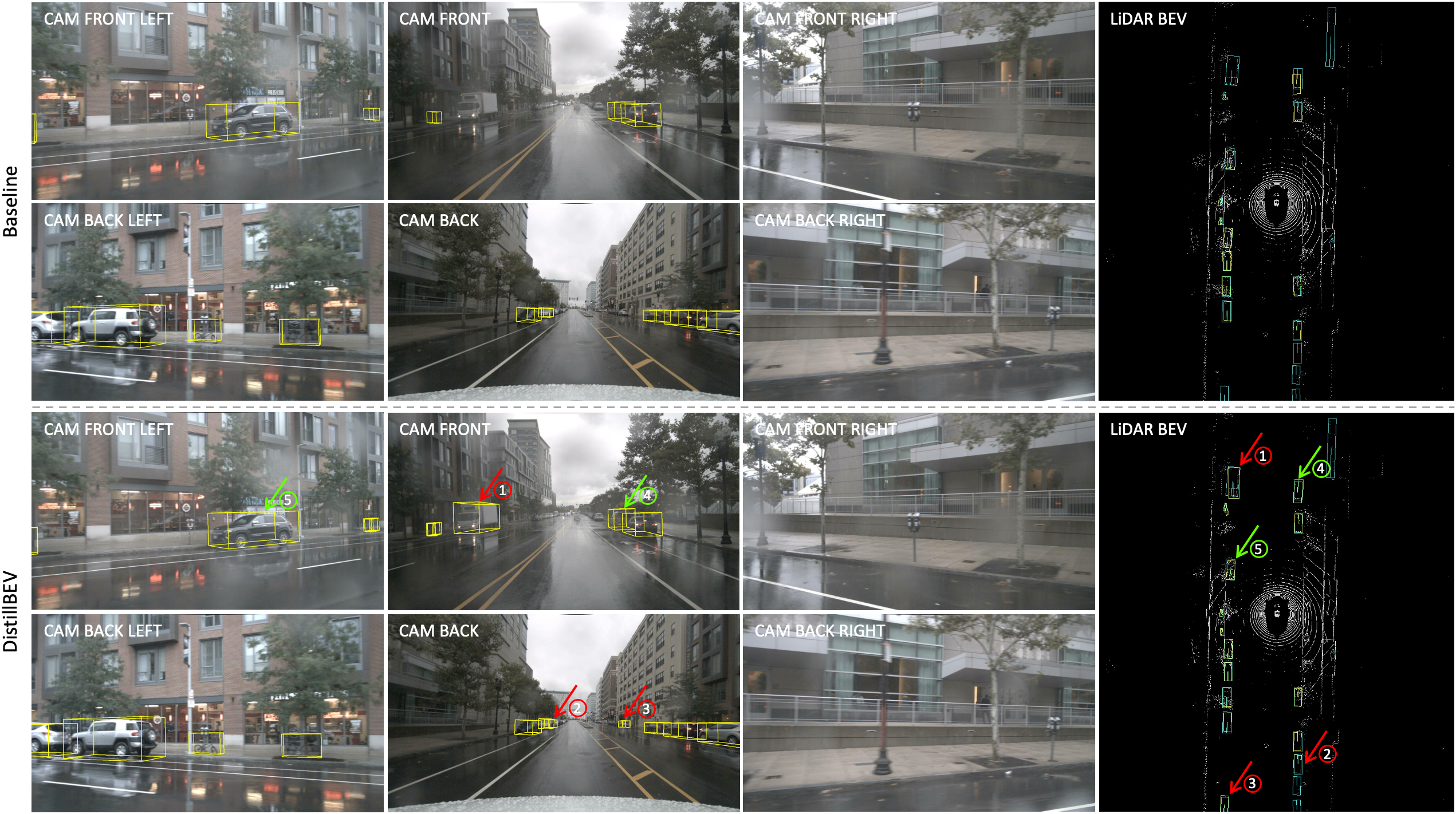}
    \caption{Comparison of the baseline (BEVDepth) and its distilled version by our approach. The cyan (in BEV only for figure clearance) and yellow boxes denote the ground truth and detection results, respectively. We use the red arrows to mark the objects that are missed by baseline but detected by DistillBEV, and the green arrows to indicate the objects that are localized more accurately by DistillBEV.}
    \label{fig:qualitative}
\end{figure*}

\begin{table*}[h]
\small
\centering
\begin{tabular}{lc|cccccccccc}
\shline
Teacher &Student &Car &Truck &Bus &Trailer &CV &Ped &Motor &Bicycle  &TC &Barrier 
 \\ \shline
BEVDepth &BEVDet & 51.4 & 23.5 & 34.0 & 14.9 & 7.1 & 33.7 & 27.8 & 23.7 & 49.2 & 49.5 \\
CenterPoint &BEVDet & \textbf{54.3} & 26.8 & 38.8 & \textbf{19.2} & 6.6 & 33.4 & 27.8 & 22.0 & 47.6 & 50.1 \\
MVP  &BEVDet & 54.2 & \textbf{28.3} & \textbf{40.4} & 16.7 & \textbf{7.4} & \textbf{35.2} & \textbf{30.2} & \textbf{25.9} & \textbf{49.7} & \textbf{52.4} \\ \shline
\end{tabular}
\vspace{5pt}
\caption{Comparison of the per-class AP on the validation set of nuScenes based on the three different teacher models to perform knowledge distillation in BEV. Abbreviations are construction vehicle (CV), pedestrian (Ped), motorcycle (Motor), and traffic cone (TC).}
\label{tab:teacher comparison mAP}
\end{table*}

\begin{table}[h]
\small
\centering
\begin{tabular}{lcc|cc}
\shline
Teacher &Mode &Params (M) &mAP &NDS \\ \shline
- &- &- &30.5 &37.8\\
\hline
BEVDepth &C & 53.2 &31.5 &38.5 \\
CenterPoint &L &6.0 &32.7 &40.7 \\
MVP &L\&C &6.0 &\textbf{34.0} &\textbf{41.6} \\ \shline
\end{tabular}
\vspace{5pt}
\caption{Comparison of three different teacher models for knowledge distillation in BEV on the validation set of nuScenes.} 
\label{tab:teacher comparison vision-based}
\end{table}

\noindent\textbf{False Positive Region Decomposition.} 
As demonstrated in Table~\ref{tab:ablation}, FP regions decomposed from background are beneficial to distillation. Built upon the optimal multi-scale setup in Table~\ref{tab:location}, we evaluate different FP region mining principles according to~(\ref{Eq:FP region}), namely the teacher-generated or student-generated FP regions under different thresholds $\gamma$. As shown in Table~\ref{tab:fp selection}, the teacher-generated FP regions with $\gamma$ = 0.1 overall performs the best. This shows that the feature responses even at FP regions produced by teacher is assistive in 3D geometric knowledge transfer. We further evaluate this strategy on multi-scale layers. As compared in Table~\ref{tab:fp selection position}, FP region decomposition performs better with the pre-head layer only. We hypothesize that such regions can be better represented by the features with high-level semantics that emerge at the last layer.  

\noindent\textbf{Adaptation Module.} 
A lightweight adaptation module is employed to map the student feature before aligning to the teacher feature at the corresponding level. This operation is necessary to match the feature size (spatial and channel dimensions) for multi-scale distillation between two discrepant architectures, and to provide extra flexibility to mitigate the difficulty in cross-modal feature imitation learning. We evaluate different design choices of the adaptation module in Table~\ref{tab:adaptation network}. 
It is observed that a simple linear projection works comparably well for the pre-head layer, while the intermediate layers require more non-linear mapping. 
This comparison suggests that the student features from intermediate layers need more adaptation in order to better align with the teacher features due to the relatively large representation gap at the early stage caused by different modalities. As for dimension matching, we can either upsample student features or downsample teacher features. Table~\ref{tab:adaptation network} shows that upsampling student features is favorable as it preserves more spatial details for feature alignment. 

\noindent\textbf{Teacher Model.}
Finally, we investigate the effect of different teacher models for knowledge distillation in BEV. Here we use BEVDet as the student model, and choose three teacher models including BEVDepth, CenterPoint and MVP, which represent the multi-camera BEV, LiDAR and camera-LiDAR fusion based detectors. Table~\ref{tab:teacher comparison vision-based} shows that the distillation across modalities clearly outperforms that within camera modality, validating the advantage of features extracted from point clouds to guide the student detector for more effective representation learning. We can further improve the student model by switching to the sensor-fusion based teacher model MVP, which shares similar network capacity as CenterPoint (both are much more compact than BEVDepth).
As shown in Table~\ref{tab:teacher comparison mAP}, compared to CenterPoint, MVP boosts the overall performance, in particular for small objects such as pedestrian and motorcycle. This is due to the fact that point clouds become increasingly sparse for small or distant objects, and augmenting point clouds with dense image information alleviates this issue.

\subsection{Qualitative Results}
In Figure~\ref{fig:qualitative}, we visualize the detection results in multi-camera images and BEV (LiDAR is provided for reference). DistillBEV successfully detects three objects that are missed by the baseline, in particular for the two distant vehicles indexed by ``2'' and ``3''. One can also see that our approach localizes objects with more accurate depth as marked by the green arrows.  
These examples qualitatively show the efficacy of the proposed approach to enhance the multi-camera BEV based 3D object detectors.  

\section{Conclusion}
We have presented DistillBEV, a simple and effective approach that guides the learning process of a multi-camera BEV based student model through a more powerful LiDAR based teacher model. Extensive experiments on eight teacher-student combinations reveal that our approach consistently renders significant performance gains with no extra computation cost during inference. This demonstrates that our approach is not just geared to a particular network but is flexible to cope with a broad range of models in both CNNs and Transformers. In the future work, we intend to explore and apply the proposed approach to more multi-camera perception tasks in BEV, such as segmentation, tracking and online high-definition map construction.  

{\small
\bibliographystyle{ieee_fullname}
\bibliography{egbib}
}

\appendix
\input{appendix.tex}

\end{document}

%% file: appendix.tex
\clearpage

\begin{figure}[h]
    \centering
    \includegraphics[width=\linewidth]{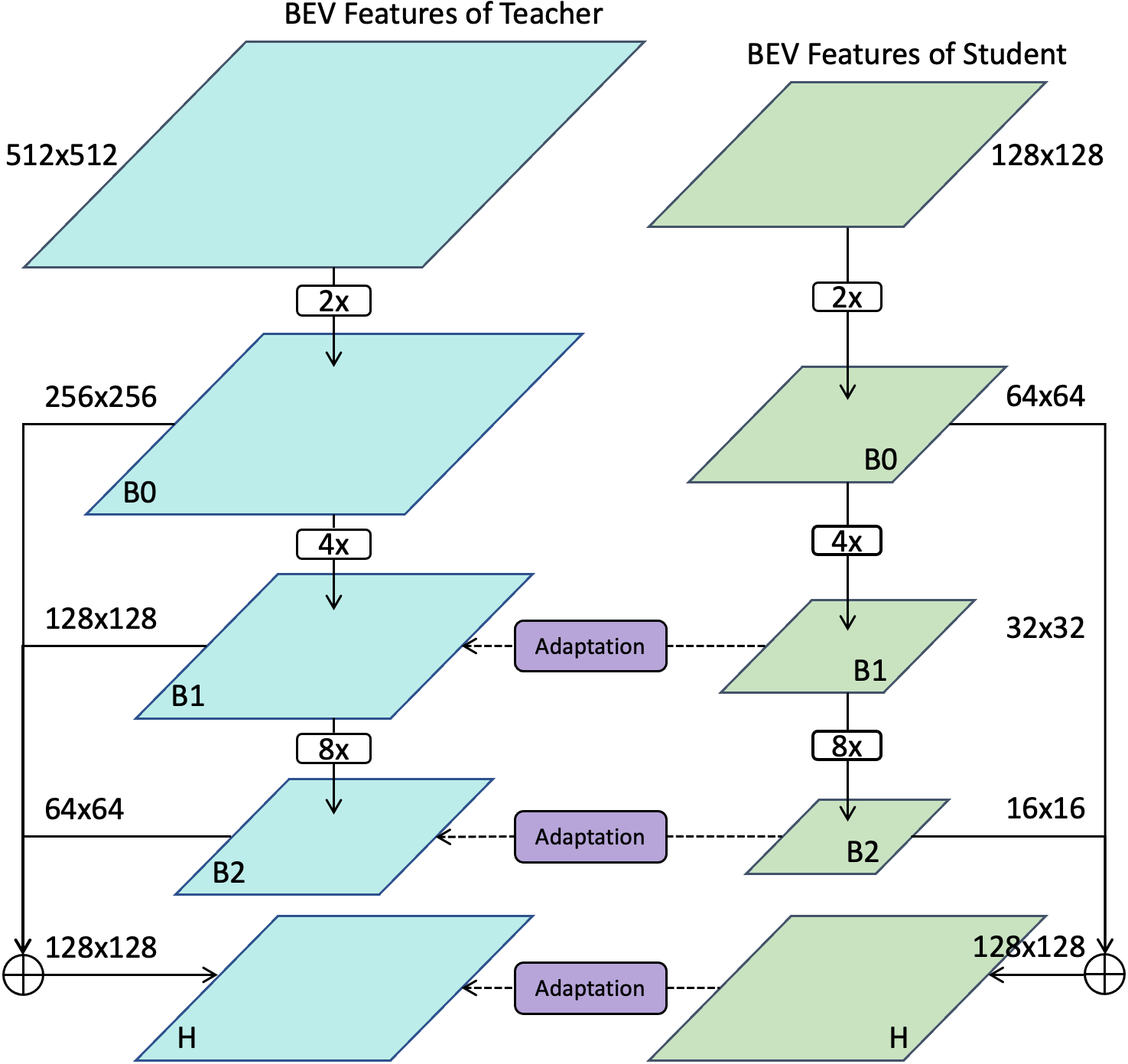}
    \caption{Illustration of the architecture details (BEV feature encoding parts) in teacher and student networks based on CNNs, as well as the multi-scale distillation performed at different levels.}
    \label{fig:arch}
\end{figure}

\section*{Appendix}

Section~\ref{sec:arch} describes the architectures of teacher and student models based on CNNs and Transformers. Section~\ref{sec:curve} reports the training process and the extra training time induced by the proposed distillation.  
Section~\ref{sec:attention} analyzes the spatial attention maps generated by both teacher and student detectors. 
Section~\ref{sec:hyper} provides more implementation details.

\section{Architectures}
\label{sec:arch}
As discussed in the main paper, the LiDAR based teacher model and the multi-camera BEV based student model are separately developed in their specific domains, resulting in different architectures. 

We first illustrate the CNNs based architectures as well as the corresponding multi-scale distillation in Figure~\ref{fig:arch}, where H indicates the pre-head layer and B2-B0 denote its three preceding intermediate layers. This figure shows the differences between teacher and student, such as feature map sizes, structures, connections, etc. We introduce the lightweight adaptation module to map the student features before aligning with the associated teacher features. It is also observed that distilling at B0 is detrimental, presumably because the representation gap between the two modalities remains large at the early stage.  

\begin{figure}[h]
    \centering
    \includegraphics[width=\linewidth]{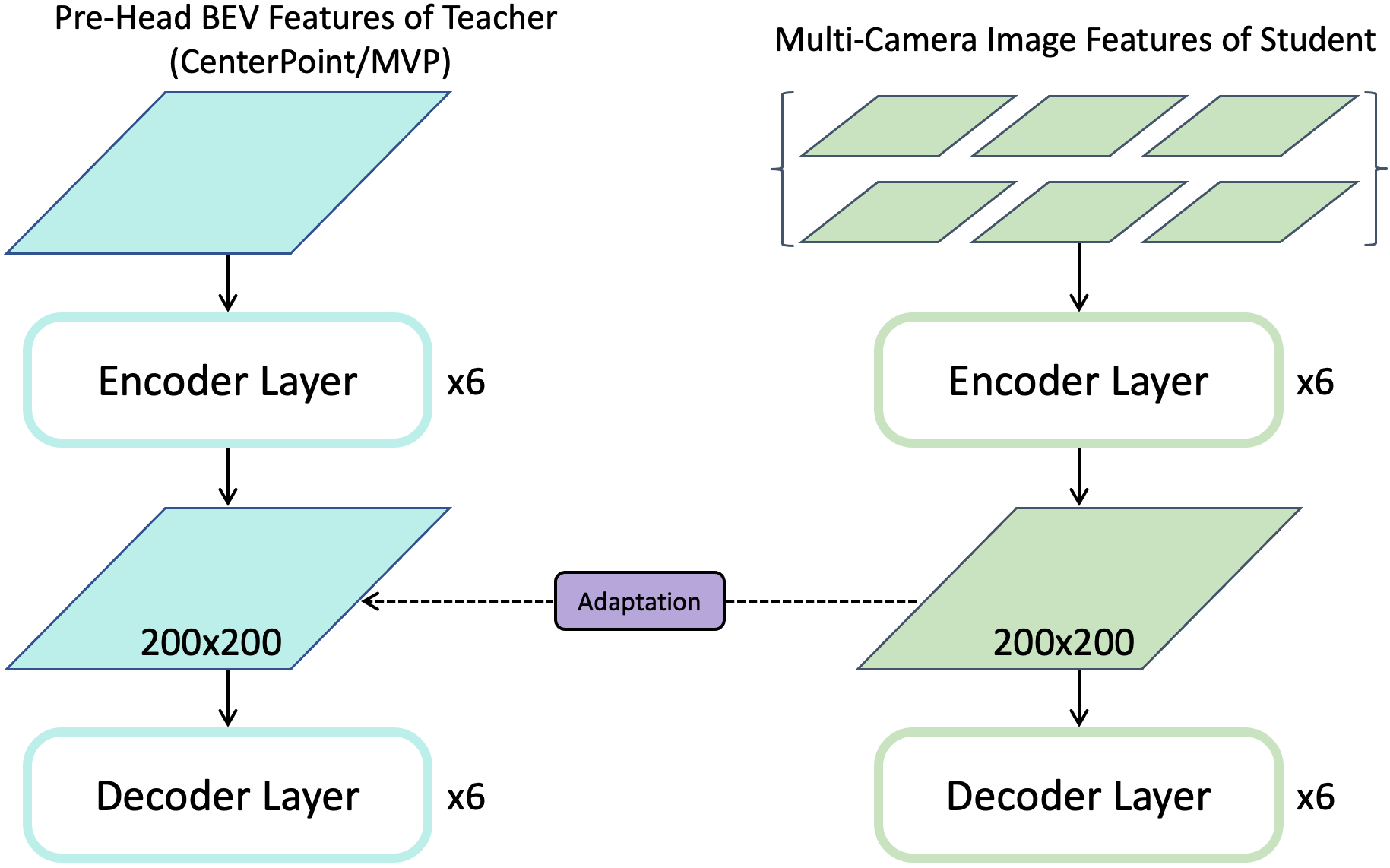}
    \caption{Illustration of the architecture details (encoder and decoder parts) in teacher and student networks based on Transformers, as well as the distillation performed at the corresponding level.}
    \label{fig:arch_transformers}
\end{figure}

To facilitate the cross-modal distillation for BEVFormer, we develop the Transformers based teacher model built on top of CenterPoint or MVP. As illustrated in Figure~\ref{fig:arch_transformers}, the distillation is performed at the intermediate features between the encoder layers and the decoder layers. 

\section{Training Process}
\label{sec:curve}
\begin{figure*}[h]
    \centering
    \includegraphics[width=\textwidth]{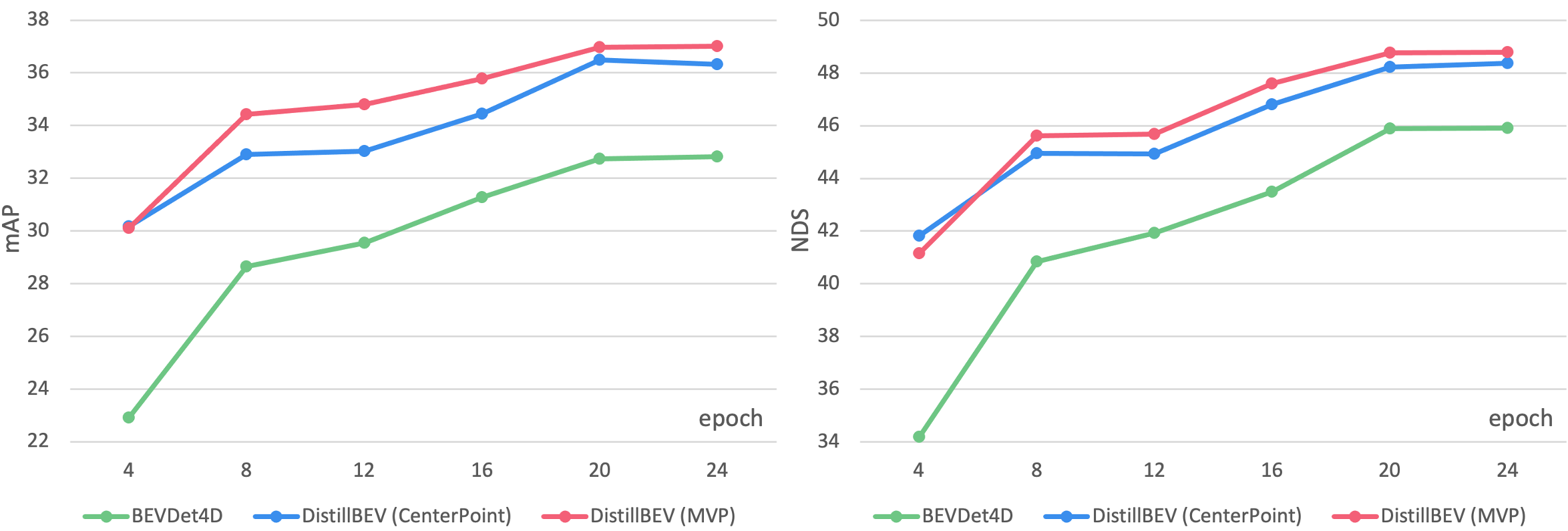}
    \caption{Comparison of the training process between the student model (BEVDet4D) and the distilled versions using CenterPoint and MVP as the teacher models. We report the results of mAP and NDS on the validation set of nuScenes.}
    \label{fig:curve}
\end{figure*}

Next we take a closer look into the learning process of the student model before and after applying DistillBEV. As shown in Figure~\ref{fig:curve}, the proposed cross-modal distillation approach brings consistent improvements over the baseline (exemplified with BEVDet4D). 
Comparing the two different teacher models, we observe that MVP (camera-LiDAR fusion) is more effective than CenterPoint (LiDAR only) to perform distillation in general, and the performance gains are not diminishing along with the training.    

As for the extra training time induced by our approach, training with 8 V100 GPUs, the student model takes 42.3 hours, and DistillBEV conducted on this model uses 49.0 hours (+15.8\%), which is a relatively low extra training cost compared to the large performance gain.

\section{Attention Visualization}
\label{sec:attention}
\begin{figure}[h]
    \centering
    \includegraphics[width=\linewidth]{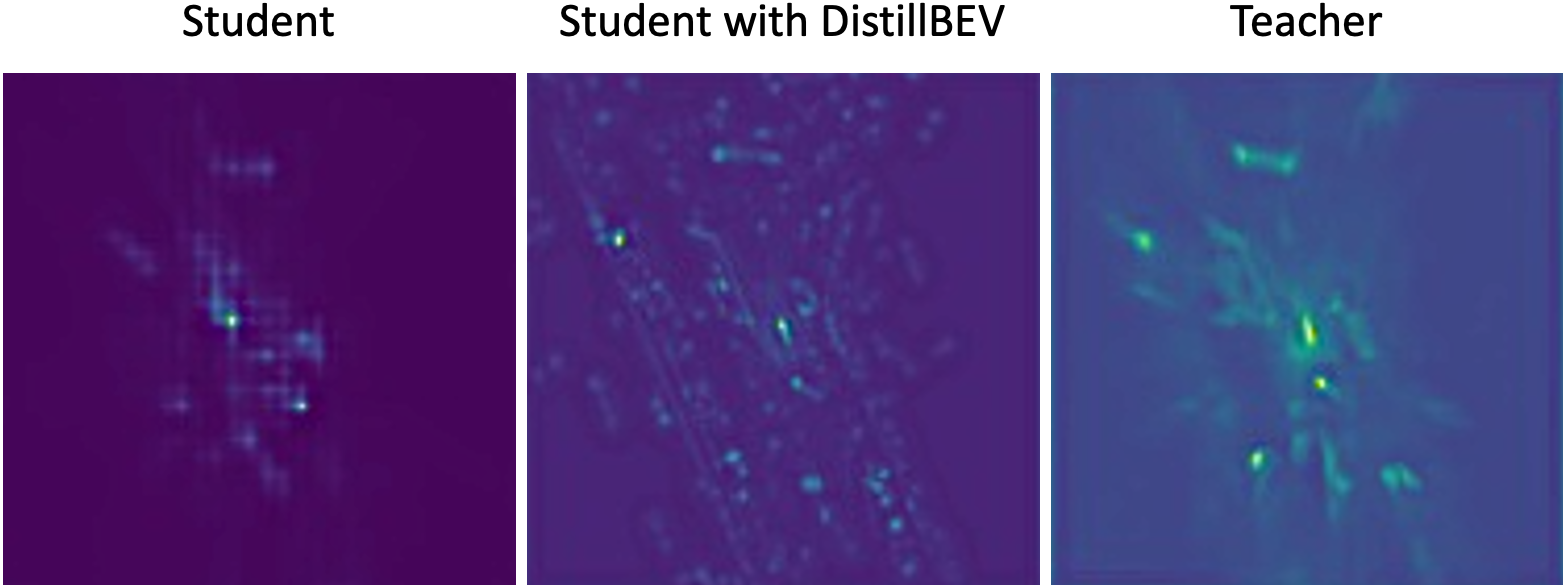}
    \caption{Visualization of the spatial attention maps generated by the teacher detector (MVP) and student detector (BEVDepth) before and after using DistillBEV.}
    \label{fig:attention}
\end{figure}

To investigate the cross-modal distillation effect to the change of student features, we visualize the spatial attention maps generated by the teacher and student models (with and without DistillBEV) following Equations (4-5) in the main paper. As shown in Figure~\ref{fig:attention}, the spatial attention map generated by the student exhibits a drastically different pattern compared to the one by the teacher model. The former concentrates on the central area (i.e., close to the ego-vehicle), and rarely activates in some distant yet important areas. After training with DistillBEV, the attention map produced by student becomes sharper and is more similar to the one of teacher in both nearby and faraway regions.  

\section{More Implementation Details}
\label{sec:hyper}
In all experiments, we set the region decomposition and spatial attention related hyper-parameters as $\eta$ = 20, $\tau$ = 0.5, $\gamma$ = 0.1. As for the loss related hyper-parameters, $\alpha$ = 6e-3, $\beta$ = 4e-2, $\lambda$ = 2.5e-3 for the networks based on CNNs, and $\alpha$ = 5e-3, $\beta$ = 4e-3, $\lambda$ = 5e-4 for the networks based on Transformers. Following tables show our approach is robust to the values of hyper-parameters $\alpha$ and $\beta$ in a wide range. Gray color indicates the default values.

\begin{table}[h]
    \centering
    \resizebox{\linewidth}{!}{
    \begin{tabular}{c|cc|cc|cc}
        \shline
        \multirow{2}{*}{$\alpha$ / $\beta$} & \multicolumn{2}{c|}{BEVDet} & \multicolumn{2}{c|}{BEVDet4D} & \multicolumn{2}{c}{BEVDepth}  \\
         & mAP & NDS & mAP & NDS & mAP & NDS  \\
        \shline
        3e-3 / 2e-2 & 32.4 & 40.9 & 35.7 & 48.2 & 38.8 & 50.3 \\
        \rowcolor{mygray1} 6e-3 / 4e-2 & 32.7 & 40.7 & 36.3 & 48.4 & 38.9 & 48.9 \\
        1.2e-2 / 8e-2 & 31.7 & 40.2 & 34.6 & 46.5 & 38.4 & 49.6 \\
        \shline
    \end{tabular}}
    \vspace{0.3mm}
    \caption{Comparison of different hyper-parameters using CenterPoint as the teacher for the students based on CNNs.}
    \label{tab:ab_center}
\end{table}

\begin{table}[h]
    \centering
    \resizebox{\linewidth}{!}{
    \begin{tabular}{c|cc|cc|cc}
        \shline
        \multirow{2}{*}{$\alpha$ / $\beta$} & \multicolumn{2}{c|}{BEVDet} & \multicolumn{2}{c|}{BEVDet4D} & \multicolumn{2}{c}{BEVDepth} \\
         & mAP & NDS & mAP & NDS & mAP & NDS \\
        \shline
        3e-3 / 2e-2 & 33.1 & 40.4 & 36.4 & 48.3 & 40.6 & 51.6 \\
        \rowcolor{mygray1} 6e-3 / 4e-2 & 34.0 & 41.6 & 37.0 & 48.8 & 40.3 & 51.0\\
        1.2e-2 / 8e-2 & 32.8 & 38.3 & 36.2 & 47.7 & 39.4 & 49.8\\
        \shline
    \end{tabular}}
    \vspace{0.3mm}
    \caption{Comparison of different hyper-parameters using MVP as the teacher for the students based on CNNs.}
    \label{tab:ab_mvp}
\end{table}

\begin{table}[h]
    \centering
    \small
    \begin{tabular}{c|cc}
        \shline
        \multirow{2}{*}{$\alpha$ / $\beta$} & \multicolumn{2}{c}{BEVFormer} \\
         & mAP & NDS \\
        \shline
         2.5e-3 / 2e-3 & 35.6 & 46.4 \\
         \rowcolor{mygray1} 5e-3 / 4e-3 & 35.6 & 47.0 \\
         1e-2 / 8e-3 & 35.0 & 46.1 \\
         \shline
    \end{tabular}
    \vspace{1.5mm}
    \caption{Comparison of different hyper-parameters using CenterPoint as the teacher for the student based on Transformers.}
    \label{tab:ab_center_former}
\end{table}

\newpage

\begin{table}[h]
    \centering
    \small
    \begin{tabular}{c|cc}
        \shline
        \multirow{2}{*}{$\alpha$ / $\beta$} & \multicolumn{2}{c}{BEVFormer} \\
         & mAP & NDS \\
        \shline
         2.5e-3 / 2e-3 & 36.1 & 46.9 \\
         \rowcolor{mygray1} 5e-3 / 4e-3 & 36.7 & 47.6 \\
         1e-2 / 8e-3 & 35.9 & 46.5 \\
         \shline
    \end{tabular}
    \vspace{1.5mm}
    \caption{Comparison of different hyper-parameters using MVP as the teacher for the student based on Transformers.}
    \label{tab:ab_mvp_former}
\end{table}

%% file: egpaper_arxiv.bbl
\begin{thebibliography}{10}\itemsep=-1pt

\bibitem{tesla}
{Tesla AI Day}.
\newblock \url{https://youtu.be/ODSJsviD_SU}, 2022.

\bibitem{planning}
Eli Bronstein, Mark Palatucci, Dominik Notz, Brandyn White, Alex Kuefler, Yiren Lu, Supratik Paul, Payam Nikdel, Paul Mougin, Hongge Chen, Justin Fu, Austin Abrams, Punit Shah, Evan Racah, Benjamin Frenkel, Shimon Whiteson, and Dragomir Anguelov.
\newblock Hierarchical model-based imitation learning for planning in autonomous driving.
\newblock In {\em IROS}, 2022.

\bibitem{caesar2020nuscenes}
Holger Caesar, Varun Bankiti, Alex~H Lang, Sourabh Vora, Venice~Erin Liong, Qiang Xu, Anush Krishnan, Yu Pan, Giancarlo Baldan, and Oscar Beijbom.
\newblock {nuScenes}: A multimodal dataset for autonomous driving.
\newblock In {\em CVPR}, 2020.

\bibitem{bev-ipm}
Yigit~Baran Can, Alexander Liniger, Danda~Pani Paudel, and Luc~Van Gool.
\newblock Structured bird's-eye-view traffic scene understanding from onboard images.
\newblock In {\em ICCV}, 2021.

\bibitem{chen2022bevdistill}
Zehui Chen, Zhenyu Li, Shiquan Zhang, Liangji Fang, Qinhong Jiang, and Feng Zhao.
\newblock {BEVDistill}: Cross-modal {BEV} distillation for multi-view {3D} object detection.
\newblock In {\em ICLR}, 2023.

\bibitem{chong2022monodistill}
Zhiyu Chong, Xinzhu Ma, Hong Zhang, Yuxin Yue, Haojie Li, Zhihui Wang, and Wanli Ouyang.
\newblock {MonoDistill}: Learning spatial features for monocular {3D} object detection.
\newblock In {\em ICLR}, 2022.

\bibitem{dai2021general}
Xing Dai, Zeren Jiang, Zhao Wu, Yiping Bao, Zhicheng Wang, Si Liu, and Erjin Zhou.
\newblock General instance distillation for object detection.
\newblock In {\em CVPR}, 2021.

\bibitem{he2016deep}
Kaiming He, Xiangyu Zhang, Shaoqing Ren, and Jian Sun.
\newblock Deep residual learning for image recognition.
\newblock In {\em CVPR}, 2016.

\bibitem{hinton2015distilling}
Geoffrey Hinton, Oriol Vinyals, and Jeff Dean.
\newblock Distilling the knowledge in a neural network.
\newblock {\em arXiv:1503.02531}, 2015.

\bibitem{hu2018squeeze}
Jie Hu, Li Shen, and Gang Sun.
\newblock Squeeze-and-excitation networks.
\newblock In {\em CVPR}, 2018.

\bibitem{huang2022bevdet4d}
Junjie Huang and Guan Huang.
\newblock {BEVDet4D}: Exploit temporal cues in multi-camera {3D} object detection.
\newblock {\em arXiv:2203.17054}, 2022.

\bibitem{huang2021bevdet}
Junjie Huang, Guan Huang, Zheng Zhu, and Dalong Du.
\newblock {BEVDet}: High-performance multi-camera {3D} object detection in bird-eye-view.
\newblock {\em arXiv:2112.11790}, 2021.

\bibitem{kang2021instance}
Zijian Kang, Peizhen Zhang, Xiangyu Zhang, Jian Sun, and Nanning Zheng.
\newblock Instance-conditional knowledge distillation for object detection.
\newblock In {\em NeurIPS}, 2021.

\bibitem{kriegeskorte2015deep}
Nikolaus Kriegeskorte.
\newblock Deep neural networks: A new framework for modelling biological vision and brain information processing.
\newblock {\em Annual Review of Vision Science}, 2015.

\bibitem{lang2019pointpillars}
Alex Lang, Sourabh Vora, Holger Caesar, Lubing Zhou, Jiong Yang, and Oscar Beijbom.
\newblock {PointPillars}: Fast encoders for object detection from point clouds.
\newblock In {\em CVPR}, 2019.

\bibitem{lee2019energy}
Youngwan Lee, Joong-won Hwang, Sangrok Lee, Yuseok Bae, and Jongyoul Park.
\newblock An energy and {GPU}-computation efficient backbone network for real-time object detection.
\newblock In {\em CVPR Workshop}, 2019.

\bibitem{li2023pillar}
Jinyu Li, Chenxu Luo, and Xiaodong Yang.
\newblock {PillarNeXt}: Rethinking network designs for {3D} object detection in {LiDAR} point clouds.
\newblock In {\em CVPR}, 2023.

\bibitem{li2023tip}
Weixin Li and Xiaodong Yang.
\newblock Transcendental idealism of planner: Evaluating perception from planning perspective for autonomous driving.
\newblock In {\em ICML}, 2023.

\bibitem{li2022unifying}
Yanwei Li, Yilun Chen, Xiaojuan Qi, Zeming Li, Jian Sun, and Jiaya Jia.
\newblock Unifying voxel-based representation with {Transformer} for {3D} object detection.
\newblock {\em NeurIPS}, 2022.

\bibitem{li2022bevdepth}
Yinhao Li, Zheng Ge, Guanyi Yu, Jinrong Yang, Zengran Wang, Yukang Shi, Jianjian Sun, and Zeming Li.
\newblock {BEVDepth}: Acquisition of reliable depth for multi-view {3D} object detection.
\newblock In {\em AAAI}, 2023.

\bibitem{li2022bevformer}
Zhiqi Li, Wenhai Wang, Hongyang Li, Enze Xie, Chonghao Sima, Tong Lu, Qiao Yu, and Jifeng Dai.
\newblock {BEVFormer}: Learning bird's-eye-view representation from multi-camera images via spatiotemporal transformers.
\newblock In {\em ECCV}, 2022.

\bibitem{lin2017feature}
Tsung-Yi Lin, Piotr Doll{\'a}r, Ross Girshick, Kaiming He, Bharath Hariharan, and Serge Belongie.
\newblock Feature pyramid networks for object detection.
\newblock In {\em CVPR}, 2017.

\bibitem{liu2022petr}
Yingfei Liu, Tiancai Wang, Xiangyu Zhang, and Jian Sun.
\newblock {PETR}: Position embedding transformation for multi-view {3D} object detection.
\newblock In {\em ECCV}, 2022.

\bibitem{loshchilov2018decoupled}
Ilya Loshchilov and Frank Hutter.
\newblock Decoupled weight decay regularization.
\newblock In {\em ICLR}, 2019.

\bibitem{luo2021simtrack}
Chenxu Luo, Xiaodong Yang, and Alan Yuille.
\newblock Exploring simple {3D} multi-object tracking for autonomous driving.
\newblock In {\em ICCV}, 2021.

\bibitem{luo2021pillar}
Chenxu Luo, Xiaodong Yang, and Alan Yuille.
\newblock Self-supervised pillar motion learning for autonomous driving.
\newblock In {\em CVPR}, 2021.

\bibitem{ma2021delving}
Xinzhu Ma, Yinmin Zhang, Dan Xu, Dongzhan Zhou, Shuai Yi, Haojie Li, and Wanli Ouyang.
\newblock Delving into localization errors for monocular {3D} object detection.
\newblock In {\em CVPR}, 2021.

\bibitem{wayformer}
Nigamaa Nayakanti, Rami Al-Rfou, Aurick Zhou, Kratarth Goel, Khaled Refaat, and Benjamin Sapp.
\newblock Wayformer: Motion forecasting via simple and efficient attention networks.
\newblock In {\em ICRA}, 2023.

\bibitem{bev-mlp}
Bowen Pan, Jiankai Sun, Ho~Yin~Tiga Leung, Alex Andonian, and Bolei Zhou.
\newblock Cross-view semantic segmentation for sensing surroundings.
\newblock {\em IEEE Robotics and Automation Letters}, 2020.

\bibitem{park2021pseudo}
Dennis Park, Rares Ambrus, Vitor Guizilini, Jie Li, and Adrien Gaidon.
\newblock Is pseudo-{LiDAR} needed for monocular {3D} object detection?
\newblock In {\em ICCV}, 2021.

\bibitem{paszke2019pytorch}
Adam Paszke, Sam Gross, Francisco Massa, Adam Lerer, James Bradbury, Gregory Chanan, Trevor Killeen, Zeming Lin, Natalia Gimelshein, and Luca Antiga.
\newblock {PyTorch}: An imperative style, high-performance deep learning library.
\newblock {\em NeurIPS}, 32, 2019.

\bibitem{philion2020lift}
Jonah Philion and Sanja Fidler.
\newblock Lift, splat, shoot: Encoding images from arbitrary camera rigs by implicitly unprojecting to {3D}.
\newblock In {\em ECCV}, 2020.

\bibitem{romero2014fitnets}
Adriana Romero, Nicolas Ballas, Samira~Ebrahimi Kahou, Antoine Chassang, Carlo Gatta, and Yoshua Bengio.
\newblock {FitNets}: Hints for thin deep nets.
\newblock {\em arXiv:1412.6550}, 2014.

\bibitem{tian2019fcos}
Zhi Tian, Chunhua Shen, Hao Chen, and Tong He.
\newblock {FCOS}: Fully convolutional one-stage object detection.
\newblock In {\em ICCV}, 2019.

\bibitem{wang2021fcos3d}
Tai Wang, Xinge Zhu, Jiangmiao Pang, and Dahua Lin.
\newblock {FCOS3D}: Fully convolutional one-stage monocular {3D} object detection.
\newblock In {\em ICCV}, 2021.

\bibitem{wang2023prophnet}
Xishun Wang, Tong Su, Fang Da, and Xiaodong Yang.
\newblock {ProphNet}: Efficient agent-centric motion forecasting with anchor-informed proposals.
\newblock In {\em CVPR}, 2023.

\bibitem{wang2022detr3d}
Yue Wang, Vitor~Campagnolo Guizilini, Tianyuan Zhang, Yilun Wang, Hang Zhao, and Justin Solomon.
\newblock {DETR3D}: {3D} object detection from multi-view images via {3D-to-2D} queries.
\newblock In {\em CoRL}, 2022.

\bibitem{wang2021object}
Yue Wang and Justin Solomon.
\newblock Object {DGCNN}: {3D} object detection using dynamic graphs.
\newblock {\em NeurIPS}, 2021.

\bibitem{yan2018second}
Yan Yan, Yuxing Mao, and Bo Li.
\newblock {SECOND}: Sparsely embedded convolutional detection.
\newblock {\em Sensors}, 2018.

\bibitem{yang2022focal}
Zhendong Yang, Zhe Li, Xiaohu Jiang, Yuan Gong, Zehuan Yuan, Danpei Zhao, and Chun Yuan.
\newblock Focal and global knowledge distillation for detectors.
\newblock In {\em CVPR}, 2022.

\bibitem{yin2021center}
Tianwei Yin, Xingyi Zhou, and Philipp Krahenbuhl.
\newblock Center-based {3D} object detection and tracking.
\newblock In {\em CVPR}, 2021.

\bibitem{yin2021multimodal}
Tianwei Yin, Xingyi Zhou, and Philipp Kr{\"a}henb{\"u}hl.
\newblock Multimodal virtual point {3D} detection.
\newblock {\em NeurIPS}, 2021.

\bibitem{zhang2021improve}
Linfeng Zhang and Kaisheng Ma.
\newblock Improve object detection with feature-based knowledge distillation: Towards accurate and efficient detectors.
\newblock In {\em ICLR}, 2021.

\bibitem{zhou2018voxelnet}
Yin Zhou and Oncel Tuzel.
\newblock {VoxelNet}: End-to-end learning for point cloud based {3D} object detection.
\newblock In {\em CVPR}, 2018.

\bibitem{zhu2019class}
Benjin Zhu, Zhengkai Jiang, Xiangxin Zhou, Zeming Li, and Gang Yu.
\newblock Class-balanced grouping and sampling for point cloud {3D} object detection.
\newblock {\em arXiv:1908.09492}, 2019.

\end{thebibliography}
